\documentclass[sigconf]{acmart}
\usepackage{courier}
\usepackage{multirow}
\usepackage{pifont}
\newcommand{\cmark}{\ding{51}}%
\newcommand{\xmark}{\ding{55}}%

\AtBeginDocument{%
  }

\setcopyright{acmcopyright}
\copyrightyear{2023}
\acmYear{2023}
\acmDOI{XXXXXXX.XXXXXXX}

\acmConference[DocIU Workshop, CIKM '23]{DocIU Workshop, CIKM '23: The 32nd ACM International Conference on Information and Knowledge Management}{October 21--25, 2023 }{Birmingham, UK}
\acmBooktitle{DocIU Workshop, CIKM '23: The 32nd ACM International Conference on Information and Knowledge Management}

\acmPrice{15.00}
\acmISBN{978-1-4503-XXXX-X/18/06}

\begin{document}

\title[MemSum-DQA: Long Document Extractive Summarizer for Document Question Answering]{MemSum-DQA: Adapting An Efficient Long Document Extractive Summarizer for Document Question Answering}


\author{Nianlong Gu}
\authornote{Corresponding author}
\affiliation{%
  \institution{Linguistic Research Infrastructure \\ University of Zurich}
  \streetaddress{Affolternstrasse 56}
  \city{Zurich}
  \country{Switzerland}
  }
\email{nianlong.gu@uzh.ch}

\author{Yingqiang Gao}
\affiliation{
  \institution{Institute of Neuroinformatics \\ University of Zurich and ETH Zurich}
  \streetaddress{Winterthurerstrasse 190}
  \city{Zurich}
  \country{Switzerland}
}
\email{yingqiang.gao@ini.ethz.ch}

\author{Richard H. R. Hahnloser}
\affiliation{
  \institution{Institute of Neuroinformatics \\ University of Zurich and ETH Zurich}
  \streetaddress{Winterthurerstrasse 190}
  \city{Zurich}
  \country{Switzerland}
}
\email{rich@ini.ethz.ch}

\renewcommand{\shortauthors}{Gu et al.}

\begin{abstract}
  We introduce MemSum-DQA, an efficient system for document question answering (DQA) that leverages MemSum, a long document extractive summarizer. By prefixing each text block in the parsed document with the provided question and question type, MemSum-DQA selectively extracts text blocks as answers from documents. On full-document answering tasks, this approach yields a 9\% improvement in exact match accuracy over prior state-of-the-art baselines. Notably, MemSum-DQA excels in addressing questions related to child-relationship understanding, underscoring the potential of extractive summarization techniques for DQA tasks.\footnote{Code is available at \url{https://github.com/nianlonggu/MemSum-DQA}}
\end{abstract}




\maketitle

\section{Introduction}
Document understanding extracts valuable information from texts, encompassing scientific papers \cite{luu2021explaining, parisot2022multi, boudin2020keyphrase, gao2022discourse}, business documents \cite{sage2020end, shukla2022dosa, agarwal2021development}, and legal documents \cite{hendrycks2021cuad, tuggener2020ledgar, nawar2022open,bauer2023legal}. Document understanding is vital for applications such as layout analysis \cite{li2020docbank}, information retrieval \cite{gu2022local}, knowledge extraction \cite{jain2020scirex}, and question answering \cite{de2019question, liu2020rikinet, zayats2021representations}.

There is a growing focus on answering questions about scientific papers, characterized by hierarchical structures such as titles, texts, lists, figures, and tables \cite{ding2023pdfvqa}. However, long document question answering is computationally challenging due to the input limitations of many pre-trained language models \cite{choi2017coarse}. Despite efforts to accommodate extended contexts in large language models \cite{beltagy2020longformer,chen2023extending}, their stringent input limits prevent the processing of some very large documents. A viable workaround could be to adapt prior work on efficient long document processing to this task.

In this study, we tackle the DocIU 2023 PDF-VQA challenge as an extractive document summarization problem rather than as a classification problem. Using MemSum \cite{gu2022memsum}, an efficient long document extractive summarizer, we pinpoint probable answer positions within documents by injecting the questions to be answered at diverse places within the document. Our proposed MemSum-DQA method achieves a 39.73\% Exact Matching Accuracy (EMA) on the test dataset of PDF-VQA (Task C, \cite{ding2023pdfvqa}). 

\begin{figure}[!t]
    \centering
    \includegraphics[width=.9\linewidth]{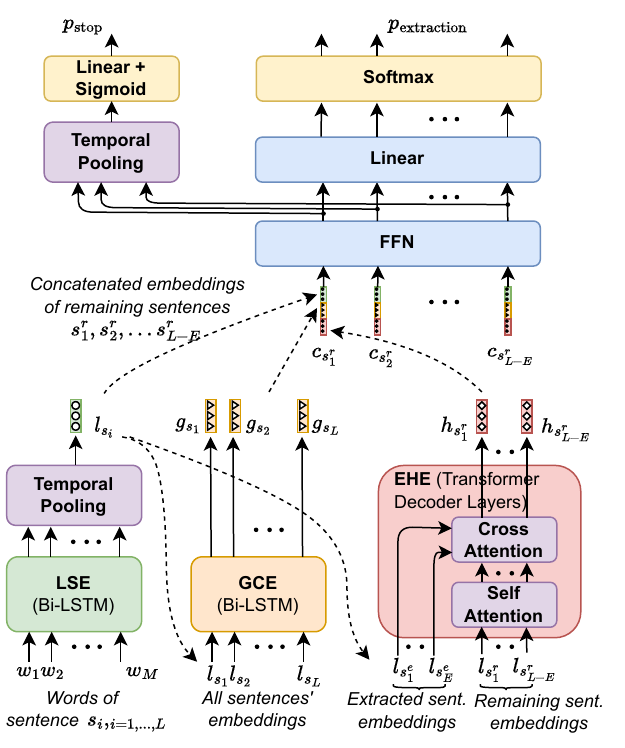}
    \caption{Schematic of MemSum-DQA.}
    \label{fig:pipeline}
\end{figure}
\begin{table*}
  \caption{Performance comparison on the PDF-VQA dataset, Task C. Features include Q (Question), B (Bounding box coordinates), V (Visual appearance), C (Context), and R (Relational Information). The relational information can be used to categorize the task (question type to be answered) into Parent- versus Child-relationship understanding, which are separately listed.  }
  \label{tab:results_comparison}
  \resizebox{.65\linewidth}{!}{ 
  \begin{tabular}{lccccccccccc}
    \toprule
    \multirow{2}*{Model} & \multicolumn{5}{c}{ Features}  &  \multicolumn{2}{c}{Parent} &  \multicolumn{2}{c}{Child} &  \multicolumn{2}{c}{Overall}   \\ \cmidrule(lr){2-6} \cmidrule(lr){7-8} \cmidrule(lr){9-10} \cmidrule(lr){11-12}
    & Q. & B. & V. & C. & R & Val & Test & Val & Test & Val & Test    \\
    \midrule
    VisualBERT \cite{li2019visualbert} & \cmark & \xmark & \cmark & \xmark & \xmark & 21.55 & 19.91 & 19.64 & 18.52 & 21.55 & 18.52 \\
    ViLT \cite{Kim2021ViLTVT} & \cmark & \xmark & \cmark & \xmark & \xmark & 11.04 & 10.21 & 8.75 & 8.79 & 10.21 & 9.87 \\
    LXMERT \cite{tan-bansal-2019-lxmert} & \cmark & \cmark & \cmark & \xmark & \xmark & 26.66 & 23.57 & 8.56 & 9.51 & 16.37 & 14.41 \\
    LoSpa \cite{ding2023pdfvqa}  & \cmark & \cmark & \cmark & \cmark & \cmark & \textbf{33.14} & 29.87 & 29.11 & 28.74 & 30.21 & 28.99 \\
    MemSum-DQA (ours) & \cmark & \xmark & \xmark & \cmark & \cmark & 32.83 & - & \textbf{73.04} & - & \textbf{40.79} & \textbf{39.73}\footnote{}  \\
  \bottomrule
\end{tabular}
}
\end{table*}
\section{Related Works}
Whereas short-document question answering has advanced thanks to improved datasets \cite{yang2015wikiqa, rajpurkar2016squad, rajpurkar2018know}, systems amenable to long documents still lag behind, as recently noted \cite{nie2023attenwalker}. 
Language models struggle when long documents surpass the input limitations of language models \cite{saad2023pdftriage}. Whereas existing workarounds pre-fetch relevant content from the entire document and represent it as plain text \cite{pereira2023visconde, gao2022precise}, this strategy often obliterates the document structure, which impacts downstream performance on document understanding tasks \cite{nie2022capturing}. Thus, to improve long document question-answering, efficient and innovative solutions are necessary to preserve document structure in an extended context for understanding tasks.
\section{Method}
\footnotetext{The test performance is assessed using approximately 40\% of the test examples, as reflected in the public Kaggle leaderboard score.}
Given parent-relationship questions such as 
``\textit{Which sections mentioned 
Fig 2?}''  or child-relationship questions such as ``\textit{Which subsections does the Results include?}'', and a PDF parsed into text blocks, the goal is to identify text block IDs that represent section or subsection names as answers (or -1 for no answer). We treat each text block as a pseudo-sentence to make the task resemble the extraction of specific sentence subsets from a document. We adapt MemSum, a model for summarizing long documents, to execute this task. Our method involves:

\noindent\textbf{Question Prefixing}: To pair up the question with possible answers, we prefix each document pseudo-sentence with the question and its type. For instance, we prefix the sentence ``\textit{An increase could be shown...}'' with the question ``\textit{Which sections mentioned Fig. 2?}'' and the type ``\textit{Parent Relationship Understanding}''.

\noindent\textbf{MemSum Adaptation} (Figure \ref{fig:pipeline}): Adhering to MemSum's framework \cite{gu2022memsum}, we perform pseudo-sentence extraction via a sequential, non-repetitive selection process. Initially, each pseudo-sentence \( s_i \) undergoes encoding through a Bi-LSTM-based Local Sentence Encoder (LSE) that computes token embeddings and consolidates these into a sentence vector \( l_{s_i} \) via a temporal pooling layer with learnable weighted averaging. Subsequently, a Global Context Encoder (GCE), also a Bi-LSTM network, processes all sentence embeddings to assimilate a document-wide context.

The Extraction History Encoder (EHE) then encodes the (pseudo-) sentence extraction history at the current step using three Transformer decoder layers \cite{vaswani2023attention}. These layers first apply self-attention on unextracted sentence embeddings, followed by cross-attention from unextracted to extracted sentence embeddings, which encapsulates information on extraction history.

For the unextracted (remaining) sentences \( s_k^r \) (suppose \( E \) out of \( L \) sentences have already been extracted), we concatenate their local \( l_{s_k^r} \), global \( g_{s_k^r} \), and extraction history embeddings \( h_{s_k^r} \), and apply a linear layer to derive logits. A softmax function then determines the extraction probability \( p_\text{extraction} \) for each sentence. Additionally, a temporal pooling layer applied to the linear layers' hidden units, followed by a linear layer and sigmoid activation, yields the extraction termination probability \( p_\text{stop} \). If \( p_\text{stop} \) surpasses a threshold of 0.2 (selected based on validation performance), extraction concludes; otherwise, the sentence with the highest \( p_\text{extraction} \) is selected, proceeding to the next iteration. This iterative process continues until either \( p_\text{stop} \) signals termination or the maximum allowable number of sentences is extracted (\( N_\text{max} \)), which we set to 4, since more than 95\% of answers comprise no more than four sentences (text blocks).

\section{Results and Discussion}
We evaluated MemSum-DQA, which focuses exclusively on text, against VisualBERT \cite{li2019visualbert} and ViLT \cite{Kim2021ViLTVT}, which rely solely on visual information, and against LXMERT \cite{tan-bansal-2019-lxmert} and LoSpa \cite{ding2023pdfvqa}, which utilize both visual and textual information. We assessed question-answering performance using the Exact Matching Accuracy (EMA) score, which corresponds to the percentage of answers that exactly match the ground truth. We separately report the performance on parent- and child-relationship understanding questions on validation and test sets.

On parent-relationship questions, MemSum-DQA's performance was comparable to that of LoSpa, whereas on child-relationship questions, MemSum-DQA's performance was largely superior (refer to Table \ref{tab:results_comparison}). We attribute this latter superiority to the LSE and the GCE, which encode text hierarchies on both section and subsection levels. Thanks to the awareness of extraction history encoded by EHE and our versatile termination criterion, MemSum-DQA tends to extract the correct set of sentences.

\section{Conclusion}
This study interprets the DocIU 2023 PDF-VQA challenge as an extractive summarization task. By prefixing questions with candidate answers and employing an adapted MemSum model, we efficiently and accurately extract relevant sentence subsets as answers. Relying solely on text, MemSum-DQA notably surpasses multi-modal baselines, excelling particularly in answering child-relationship understanding questions. Future endeavors may focus on enhancing MemSum-DQA through visual information integration and exploring advanced techniques for sentence contextualization beyond straightforward question prefixing.


\bibliographystyle{ACM-Reference-Format}
\bibliography{sample-base}

\end{document}